\documentclass[11pt,journal,compsoc]{IEEEtran}

\usepackage[cmex10]{amsmath}
\usepackage{amssymb,latexsym,xcolor,amsfonts,undertilde,cancel,mathabx}
\usepackage{graphicx,MnSymbol}
\usepackage{wrapfig,subfig,enumerate}

\usepackage{amsthm}
\theoremstyle{plain}

\theoremstyle{definition}
\newtheorem*{definition}{Definition}
\theoremstyle{remark}

\newtheorem*{example}{Example}

\newcommand{\R}{\mathbb{R}}

\newcommand{\C}{\mathbb{C}}

\newcommand{\beq}{\begin{equation}}
\newcommand{\eeq}{\end{equation}}
\newcommand{\ba}{\begin{array}}
\newcommand{\ea}{\end{array}}
\newcommand{\bea}{\begin{eqnarray}}
\newcommand{\eea}{\end{eqnarray}}
\newcommand{\bc}{\begin{center}}
\newcommand{\ec}{\end{center}}

\newcommand{\bt}{\begin{table}}
\newcommand{\et}{\end{table}}
\newcommand{\la}[1]{\label{#1}}

\newcommand{\D}[1]{\Delta #1}

\newcommand{\beqs}{\begin{equation*}}
\newcommand{\eeqs}{\end{equation*}}
\newcommand{\aln}{\begin{aligned}}
\newcommand{\ealn}{\end{aligned}}
\newcommand{\bcs}{\begin{cases}}
\newcommand{\ecs}{\end{cases}}
\newcommand{\bmat}{\begin{bmatrix}}
\newcommand{\ebmat}{\end{bmatrix}}

\newcommand{\ben}{\begin{enumerate}}
\newcommand{\een}{\end{enumerate}}

\newcommand{\jth}[1]{#1^{\text{th}}}

\newcommand{\bfvarphi}{\boldsymbol\varphi}
\newcommand{\bfPhi}{\boldsymbol\Phi}

\newcommand{\bfOmega}{\boldsymbol\Omega}
\definecolor{dgreen}{rgb}{0.2,0.5,0.2}
\definecolor{gold}{rgb}{0.8, 0.498, 0.196}
\definecolor{purple}{rgb}{0.502, 0, 0.502}
\definecolor{dorange}{rgb}{1,.549,0}
\definecolor{dorchid}{rgb}{0.6,.196,0.8}
\definecolor{gold}{rgb}{0.8,0.498,0.196}

\begin{document}


\title{Dynamic Mode Decomposition for Real-Time Background/Foreground Separation in Video}

\author{J.~Grosek and J.~Nathan~Kutz
\thanks{ J. Grosek and J. N. Kutz are with the Department of Applied Mathematics, University of Washington, Box 353925, Seattle, WA, 98195-3925}
\thanks{J. Grosek acknowledges support from the SMART Scholarship Program.  J. N. Kutz acknowledges support from the National Science Foundation (NSF)  (DMS-1007621) and the US Air Force Office of Scientific Research (AFOSR) (FA9550-09-0174).}
\thanks{J. Grosek:  jgrosek@uw.edu,  J. N. Kutz: kutz@uw.edu}}

\IEEEcompsoctitleabstractindextext{
\begin{abstract}
This paper introduces the method of dynamic mode decomposition (DMD) for robustly separating video frames into background (low-rank) and foreground (sparse) components in real-time.  
The method is a novel application of a technique used for characterizing nonlinear dynamical systems in an equation-free manner by decomposing the state of the system into low-rank terms whose Fourier components in time are known.  
DMD terms with Fourier frequencies near the origin (zero-modes) are interpreted as background (low-rank) portions of the given video frames, and the terms with Fourier frequencies bounded away from the origin are their sparse counterparts.  
An approximate low-rank/sparse separation is achieved at the computational cost of just one singular value decomposition and one linear equation solve, thus producing results orders of magnitude faster than a leading separation method, namely robust principal component analysis (RPCA).  
The DMD method that is developed here is demonstrated to work robustly in real-time with personal laptop-class computing power and without any parameter tuning, which is a transformative improvement in performance that is ideal for video surveillance and recognition applications.  
\end{abstract}}

\maketitle

\IEEEdisplaynotcompsoctitleabstractindextext
\IEEEpeerreviewmaketitle

\newsavebox{\matS}
\savebox{\matS}{$
\bmat
	0 & 0 & \cdots & 0 & 0 & k_{1} \\
	1 & 0 & \cdots & 0 & 0 & k_{2} \\
	0 & 1 & \ddots & 0 & 0 & k_{3} \\
	\vdots & \vdots & \ddots & \vdots & \vdots & \vdots \\
	0 & 0 & \cdots & 1 & 0 & k_{m - 2} \\
	0 & 0 & \cdots & 0 & 1 & k_{m - 1} \\
\ebmat
$}
\newsavebox{\matOmega}
\savebox{\matOmega}{$
\bmat
	e^{\omega_{1}} & 0 & \cdots & 0 \\
	0 & e^{\omega_{2}} & \ddots & 0 \\
	\vdots & \ddots & \ddots & \vdots \\
	0 & 0 & \cdots & e^{\omega_{\ell}} \\
\ebmat
$}



\section{Introduction}\la{sect:Intro}

There is a growing demand for accurate and real-time video surveillance techniques.  Specifically, algorithms that can remove {\em background} variations in a video stream, which are highly correlated between frames, in order to highlight {\em foreground} objects of potential interest are at the forefront of modern data-analysis research.  Background/foreground separation is typically an integral step in detecting, identifying, tracking, and recognizing objects in video sequences.  Most modern computer vision applications demand algorithms that can be implemented in real-time, and that are robust enough to handle diverse, complicated, and cluttered backgrounds.  Competitive methods often need to be flexible enough to accommodate changes in a scene due to, for instance, illumination changes that can occur throughout the day, or location changes where the application is being implemented.  
Given the importance of this task, a variety of iterative techniques and methods, some of which are outlined below, have already been developed in order to perform background/foreground separation~\cite{StatSep,Gaussian,NeuralNet,GRASTA,RPCA1} (See also, for instance, the recent review~\cite{back_review} which compares error and timing of various methods).  

One potential viewpoint of this computational task is as a matrix separation problem into {\em low-rank} (background) and {\em sparse} (foreground) components.  Recently, this viewpoint has been advocated by Cand\`{e}s et al. in the framework of {\em robust principal component analysis} (RPCA)~\cite{RPCA1}.  By weighting a combination of the nuclear and the $L^{1}$ norms, a convenient convex optimization problem ({\em principal component pursuit}) was demonstrated, under suitable assumptions, to recover the low-rank and sparse components exactly of a given data-matrix (or video for our purposes).  
The RPCA technique, which has its computational costs dominated by the convex optimization procedure, was shown to be highly-competitive in comparison to the state-of-the-art computer vision procedure developed by De La Torre and Black~\cite{RPCA2}.  In this manuscript, we advocate a similar matrix separation approach, but by using the method of {\em dynamic mode decomposition} (DMD)~\cite{DMD1,DMD2,DMD3,DMD4,DMD5,DataBook} instead of RPCA~\cite{RPCA1}.  This method, which essentially implements a Fourier decomposition of the video frames in time, distinguishes the stationary background from the dynamic foreground by differentiating between the near-zero modes and the remaining modes bounded away from the origin, respectively.  

Originally introduced in the fluid mechanics community, DMD has emerged as a powerful tool for analyzing the dynamics of nonlinear systems~\cite{DMD1,DMD2,DMD3,DMD4,DMD5,DataBook}.  
In the context of fluids, DMD has gained popularity since it provides, in an {\em equation-free} manner, information about the dynamics of flow even if the underlying dynamics are nonlinear.  It is equation-free in the sense that a typical application requires collecting a time series of experimental (or simulated) velocity fields, and computing DMD modes and eigenvalues from them.  The modes are spatial fields that often identify coherent structures in the flow.  The corresponding eigenvalues define growth/decay rates and oscillation frequencies for each mode.  
Taken together, the DMD modes and eigenvalues describe the dynamics observed in the time series in terms of growth, decay, and oscillatory components; i.e. it is a decomposition of the data into Fourier modes in time~\cite{DMD3}.  

In the application of video surveillance, the video frames can be thought of as snapshots of some underlying complex/nonlinear dynamics.  
The DMD decomposition yields oscillatory time components of the video frames that have contextual implications.  Namely, those modes that are near the origin represent dynamics that are unchanging, or changing slowly, and can be interpreted as stationary background pixels, or low-rank components of the data matrix.  
In contrast, those modes bounded away from the origin are changing on $\mathcal{O}(1)$ timescales or faster, and represent the foreground motion in the video, or the sparse components of the data matrix.  
Thus, by simply applying the dynamical systems DMD interpretation to video frames, an approximate RPCA technique can be enacted at a fixed cost of a singular-value decomposition and a linear equation solve.  Unlike the convex optimization procedure of Cand\`{e}s et al.~\cite{RPCA1}, which can be guaranteed to exactly produce a low-rank and sparse separation under certain assumptions, no such guarantees are currently given for the DMD procedure.  Regardless, in comparison with the RPCA~\cite{RPCA1} and computer vision~\cite{RPCA2} methods, the DMD procedure is orders of magnitude faster in computational performance, resulting in real-time separation on laptop-class computing power, which suggests that this DMD technique merits serious consideration.

This paper is organized as follows: 
First, a brief synopsis of RPCA theory is given, with an emphasis on its interpretation in the computer vision field.  
Next, a complete review of DMD theory is outlined, culminating in the methodology that allows the DMD method to separate an approximate low-rank structure from a sparse structure, similar to the RPCA method.  
Then the DMD separation algorithm is tested on an example where noise needs to be removed from a sampled solution of a dynamical system.  
Finally, the DMD and RPCA techniques are compared and contrasted as they are applied to real surveillance video footage and to a constructed video used for error analysis, where moving objects will be extracted from their background.  The performance and limitations of the DMD method are discussed as well.  In the conclusion, the merits of the DMD method for background/foreground will be summarized, and some future and open projects will be suggested.  


\section{Robust PCA Theory}\la{sect:RPCAT}

For many modern data analysis problems, {\em principal component analysis} (PCA) has become a fairly ubiquitous means for achieving dimensionality reduction.  
The PCA method, and many other statistical methods for reducing dimensionality, rely on the $L^{2}$-norm for optimizing data-fitting because of its convenient mathematical properties and ease of interpretation as a measure of an {\em energy} in the system.  
However, the $L^{2}$-norm suffers greatly in the presence of corrupted data because of its sensitivity to outliers, which easily skew the results by weighing them more heavily than one would desire.  

There are other norms available to be used as measures for data-fitting; specifically, the $L^{1}$-norm is a good candidate because it does not over-weight sparse outliers in the data and can be used to promote sparsity.  
Indeed, the $L^{1}$-norm has been recently proved to be a broad measure with many favorable physical interpretations for various applications~\cite{RPCA8}.  
For the specific task considered here, the RPCA algorithm improves the classical PCA algorithm by mixing the use of the two different norms in order to make the dimensionality reduction more robust to potentially corrupt and/or sparse data.  

\subsection{RPCA Algorithm}\la{subsect:RPCAA}

Given a collection of data from a potentially complex, nonlinear system, the RPCA method will seek out the sparse structures within the data, while simultaneously fitting the remaining entries to a low-rank basis.  
As long as the given data is truly of this nature, in that it lies on a low-dimensional subspace and has sparse components, then the RPCA algorithmhas been proven by Cand\`{e}s et al.~\cite{RPCA1} to perfectly separate the given data ${\bf X}$ according to ${\bf X} = {\bf L} + {\bf S}$, where ${\bf L}$ has a low-rank and ${\bf S}$ is sparse.  
 
The key to the RPCA algorithm is formulating the problem into a tractable, nonsmooth convex optimization problem known as {\em principal component pursuit} (PCP):
\beq\la{equ:PCP}
	\aln
		\arg \min & \ \ \|{\bf L}\|_{*} + \lambda\|{\bf S}\|_{1} \\
		\text{subject to} & \ \ {\bf X} = {\bf L} + {\bf S}
	\ealn
\eeq
Here PCP is minimizing the weighed combination of the nuclear norm: $\|{\bf M}\|_{*} := \text{trace}\left(\sqrt{{\bf M}^{*}{\bf M}}\right)$ and the $L^{1}$-norm: $\|{\bf M}\|_{1} := \sum_{ij}|m_{ij}|$.  
The scalar regularization parameter is nonnegative: $\lambda \geq 0$.  
From the optimization problem \eqref{equ:PCP}, it can be seen that as $\lambda \to 0$, the low-rank structure will incorporate all of the given data: ${\bf L} \to {\bf X}$, leaving the sparse structure devoid of anything.  It is also true that as $\lambda$ increases, the sparse structure will embody more and more of the original data matrix: ${\bf S} \to {\bf X}$, as the low-rank structure will commensurately approach the zero matrix~\cite{RPCA1,DataBook}.  

Effectively, $\lambda$ controls the dimensionality of the low-rank subspace; however, one does not need to know the rank of ${\bf L}$ {\em a priori}.  Cand\`{e}s et al.~\cite{RPCA1} have shown that the choice
\[
	\lambda = \frac{1}{\sqrt{\max(n,m)}},
\]
where ${\bf X}$ is $n \times m$, has a high probability of success at producing the correct low-rank and sparse separation provided that the matrices ${\bf L}$ and ${\bf S}$ are incoherent, which is the case for many practical applications.  Some fine tuning of $\lambda$ may yield slightly improved results~\cite{DataBook}.  

Though there are multiple methods that can solve the convex PCP problem, the {\em augmented Lagrange multiplier} (ALM) method stands out as a simple and stable algorithm with robust, efficient performance characteristics.  
The ALM method is effective because it achieves high accuracies in fewer iterations when compared against other competing methods~\cite{RPCA1}.  
Moreover, there is an {\em inexact} ALM variant~\cite{InexactALM} to the {\em exact} ALM method~\cite{ExactALM}, which is able to converge in even fewer iterations at the cost of weaker guaranteed convergence criteria.  
Matlab code that implements these methods, along with a few other algorithms, can be downloaded from the University of Illinois Perception and Decision Lab website~\cite{ALMwebsite}, and is the code implemented throughout this manuscript for comparison purposes.  

\subsection{Video Interpretation of the RPCA Method}\la{subsect:VIRPCAM}

In a video sequence, stationary background objects translate into highly correlated pixel regions from one frame to the next, which suggests a low-rank structure within the video data.  In the case of videos, where the data in each frame is 2D by nature, frames need to be reshaped into 1D column vectors and united into a single data matrix ${\bf X}$.  
The RPCA algorithm can then implement the background/foreground separation found in ${\bf X} = {\bf L} + {\bf S}$, where the low-rank matrix ${\bf L}$ will render the video of just the background, and the sparse matrix ${\bf S}$ will render the complementary video of the moving foreground objects.  
Because the foreground objects exhibit a spatial coherency throughout the video, the RPCA method is no longer guaranteed a high probability of success; however, in practice, RPCA achieves an acceptable separation almost every time~\cite{RPCA1}.  

Figure \ref{fig:LambdaTest} illustrates the quality of background/foreground separation as the regularization parameter $\lambda$ is varied about its suggested value of $\lambda = (\sqrt{n})^{-1}$.  
In this case $n = 11520$, being the number of pixels per frame; hence, $\lambda \approx 9.32 \cdot 10^{-3}$.  
The RPCA method performs the foreground/background separation with a quality on par with the DMD method, which will be reviewed in the next section.  
However, as $\lambda$ is decreased, the sparse reconstruction of the video, which is stored in matrix ${\bf S}$, starts to bring in more of the original video, including erroneous stationary pixels that should be part of the low-rank background.  
When $\lambda$ is increased, the sparse reconstruction of the video begins to see a decrease in the pixel intensities that correspond to the moving objects, and some foreground pixels disappear all together.  

\begin{figure}[t!]
	\bc
   		\includegraphics[width=0.8\linewidth]{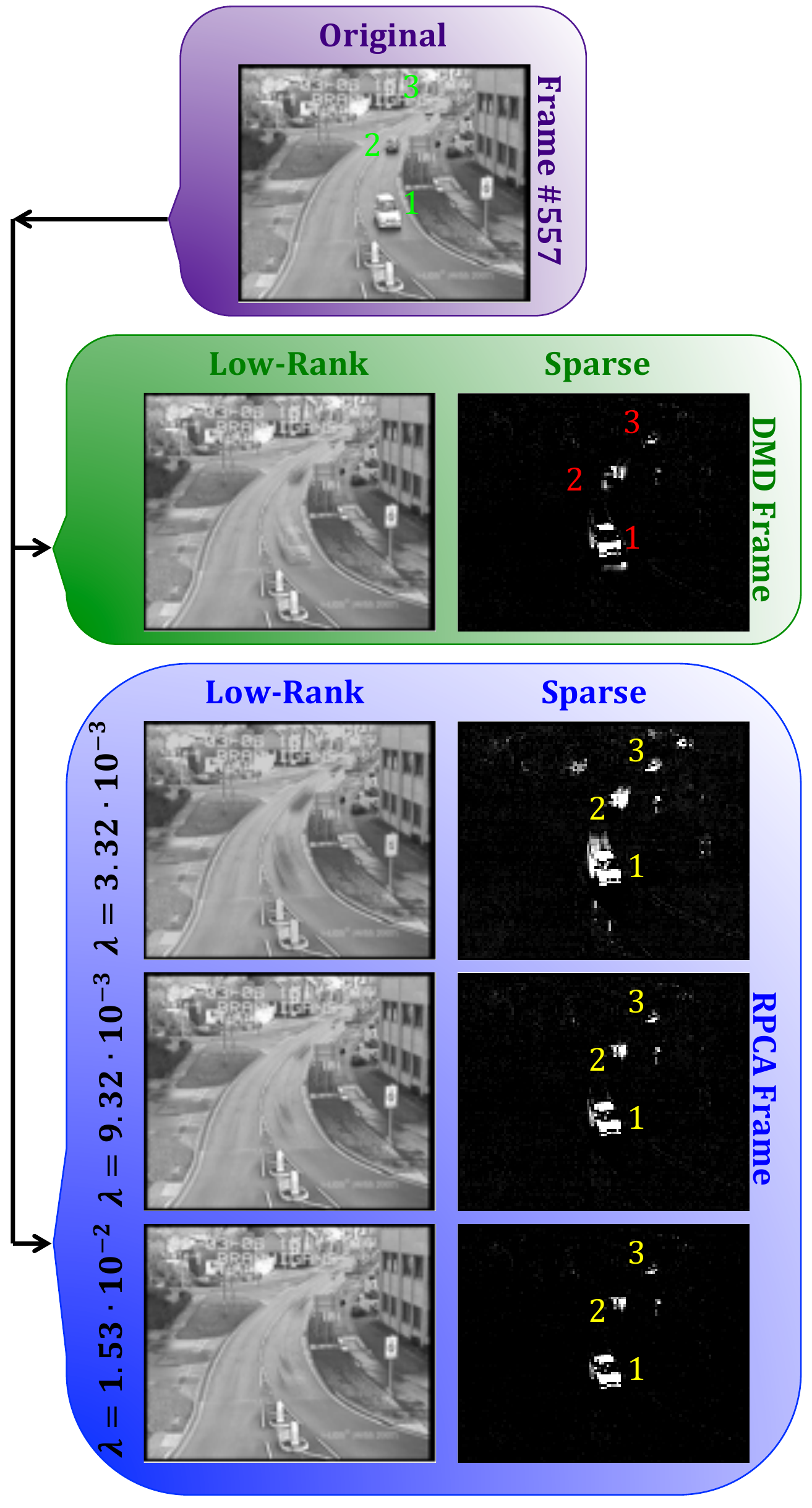}
	\ec
 	\caption{This figure demonstrates the sensitivity of the RPCA method to the $\lambda$ parameter value.  
	In Frame $\#557$ of the ``Parked Vehicle'' video~\cite{AVSS}, there are 3 moving vehicles to be separated from the background.  
	Using video segments of 30 frames, the DMD background/foreground separation method finds the 3 moving vehicles, with few erroneous pixels included as part of the foreground.  
	The RPCA method also achieves similar results with the recommended regularization parameter value of $\lambda = 9.32 \cdot 10^{-3}$.  
	As this parameter is changed by only $\pm0.006$, either extra, foreign pixels get included in the foreground ($\lambda = 3.32 \cdot 10^{-3}$), or two of the three cars begin to fade away and blend into the background ($\lambda = 1.532 \cdot 10^{-2}$).  
	Note that the sparse results for both the DMD and RPCA methods are artificially brightened by a factor of 10 in order to illuminate all the dark extraneous pixels that would normally not be visible against the black background.}
	\la{fig:LambdaTest}
\end{figure}


\section{DMD Theory}\la{sect:DMDT}

Dynamic mode decomposition (DMD) is a mathematical method that was developed in order to understand, control, or simulate inherently complex, nonlinear systems without necessarily knowing  fully, or partially, the underlying governing equations that drive the system.  
Experimental (or simulated) data, collected in snapshots through time, can be processed with DMD in order to mimic, control, or analyze the current state and/or dimensionality of a system, or even to predict future states of, or find coherent structures within, that system.  The power of DMD is found in exploiting the intrinsic low-dimensionality of a complicated system, which may not be obvious {\em a priori}, and then rendering that system in a more computationally and theoretically tractable (low-dimensional) form.  DMD has been traditionally used in the fluid mechanics, atmospheric science, and nonlinear waves communities as a popular method for data-base learning and discovery~\cite{DMD1,DMD2,DMD4,DataBook}.  

\subsection{DMD Algorithm}\la{subsect:DMDA}

Given data that is collected in regularly spaced time intervals, the DMD method will approximate the low-dimensional modes of the linear, time-independent {\em Koopman operator} in order to estimate the potentially nonlinear dynamics of the system.  This process is distinct from linearizing the dynamics.  The dynamic information is resolved through a similar process as is found in the {\em Arnoldi algorithm}~\cite{tre,DataBook}.

Consider $n$ data points collected at a given time, with a total of $m$ samplings in time, evenly spaced by $\D{t}$.  
Let ${\bf x}_{j}$ be a vector of the $n$ data points collected at time $t_{j}, \ j = 1, 2, \ldots, m$.  
The data can be grouped into matrices as follows\dots
\begin{align*}
	{\bf X}_{1}^{m - 1} & = \Big[ {\bf x}_{1} \ {\bf x}_{2} \ {\bf x}_{3} \ \ldots \ {\bf x}_{m - 1} \Big] \\
	{\bf X}_{2}^{m} & = \Big[ {\bf x}_{2} \ {\bf x}_{3} \ {\bf x}_{4} \ \ldots \ {\bf x}_{m} \Big]
\end{align*}
The Koopman operator ${\bf A}$ maps the data at time $j$ to the data at time $j + 1$ such that ${\bf x}_{j + 1} = {\bf A}{\bf x}_{j}$.  
The DMD algorithm will estimate the Koopman operator ${\bf A}$ that best represents the data in ${\bf X}_{1}^{m - 1}$ such that the columns of
\beqs
	{\bf X}_{1}^{m - 1} = \Big[ {\bf x}_{1} \ {\bf A}{\bf x}_{1} \ {\bf A}^{2}{\bf x}_{1} \ \ldots \ {\bf A}^{m - 2}{\bf x}_{1} \Big]
\eeqs
form a Krylov space.  
Thus, ${\bf A}{\bf X}_{1}^{m - 1} = {\bf X}_{2}^{m}$.  
The last data point ${\bf x}_{m}$ is determined by 
\beqs
	{\bf x}_{m} = \sum_{j = 1}^{m - 1} k_{j}{\bf x}_{j} + {\bf r},
\eeqs
where the $k_{j}$'s are the coefficients of the Krylov space basis vectors and the residual error ${\bf r}$ is orthogonal to the Krylov space~\cite{DataBook}.  

Notice that ${\bf X}_{2}^{m} = {\bf X}_{1}^{m - 1}{\bf S} + {\bf r} \cdot {\bf e}^{*}_{m - 1}$, where $[\mbox{}^{*}]$ is the conjugate transpose operator, ${\bf e}_{m - 1}$ is the $\jth{(m - 1)}$ unit vector, and 
\beqs
	{\bf S} = \usebox{\matS}
\eeqs
is a $n \times (m - 1)$ matrix.  
Also, note that the {\em singular value decomposition} (SVD) of the matrix ${\bf X}_{1}^{m - 1}$ can be used for dimensionality reduction, and for the ease and convenience of working with unitary and diagonal matrices.  
Thus, ${\bf X}_{1}^{m - 1} = {\bf U}{\bf \Sigma}{\bf V}^{*}$, where ${\bf U} \in \C^{n \times \ell}$ is unitary, ${\bf \Sigma} \in \C^{\ell \times \ell}$ is diagonal, and ${\bf V} \in \C^{(m - 1) \times \ell}$ is unitary.  
The parameter $\ell$ is chosen so as to reduce rank of ${\bf X}_{1}^{m - 1}$ as much as possible, while still capturing the fundamental structure and dynamics of the system represented by the data in ${\bf X}_{1}^{m - 1}$.  
Thus, assuming that the residual error is small ($\|{\bf r}\| \ll 1$), one can estimate
\begin{align*}
	{\bf X}_{2}^{m} & \approx {\bf X}_{1}^{m - 1}{\bf S} \\
	{\bf X}_{2}^{m} & \approx {\bf U}{\bf \Sigma}{\bf V}^{*}{\bf S} \\
	{\bf S} & \approx {\bf V}{\bf \Sigma}^{-1}{\bf U}^{*}{\bf X}_{2}^{m}.
\end{align*}
Using the similarity transform ${\bf V}{\bf \Sigma}^{-1}$, the matrix
\[
	{\bf \tilde{S}} \approx {\bf U}^{*}{\bf X}_{2}^{m}{\bf V}{\bf \Sigma}^{-1}
\]
can be derived, which is mathematically similar to the matrix ${\bf S}$~\cite{DataBook}.  

Essential to the DMD method is the idea that because ${\bf A}{\bf X}_{1}^{m - 1} = {\bf X}_{2}^{m} \approx {\bf X}_{1}^{m - 1}{\bf S}$, then some of the eigenvalues of the matrix ${\bf S}$ approximate the eigenvalues of the Koopman operator ${\bf A}$, similar to the calculation done in the Arnoldi algorithm to get the {\em Ritz values}.  
This fact can be seen in the similarity relation ${\bf A}{\bf U} \approx {\bf U}{\bf \tilde{S}}$.  
The Arnoldi iteration solves ${\bf A}{\bf Q} = {\bf Q}{\bf H}$, where ${\bf Q}$ is normal and ${\bf H}$ is an upper Hessenberg matrix.  
The connection between the Arnoldi algorithm and DMD is found by performing the {\em QR-factorization} ${\bf X}_{1}^{m - 1} = {\bf Q}{\bf R}$ and by using the relation ${\bf H} = {\bf R}{\bf S}{\bf R}^{-1}$~\cite{DMD1}.  

In practice, the eigenvalues are determined through the matrix ${\bf \tilde{S}}$ because it is similar to the matrix ${\bf S}$, and therefore is also similar to the Koopman operator ${\bf A}$.  
However, the eigenvectors of matrix ${\bf \tilde{S}}$ also approximate those of {\bf A} since
\begin{align*}
	{\bf A}{\bf U} & \approx {\bf U}{\bf \tilde{S}} = {\bf U}{\bf W}\bfOmega{\bf W}^{-1} \\
	{\bf A}\left({\bf U}{\bf W}\right) & \approx \left({\bf U}{\bf W}\right)\bfOmega \\
	{\bf A}\bfPhi & \approx \bfPhi\bfOmega,
\end{align*}
where the eigen-decomposition of ${\bf \tilde{S}}$ is ${\bf \tilde{S}} = {\bf W}\bfOmega{\bf W}^{-1}$ and $\bfPhi := {\bf U}{\bf W}$.  
Thus, solving the eigenvalue problem: 
\beq\la{equ:DMDEigenvalueProblem}
	{\bf \tilde{S}}{\bf w}_{j} = \mu_{j}{\bf w}_{j}, \ j = 1, 2, \ldots, \ell
\eeq
will yield the DMD eigenvalues $\mu_{j}$ and the eigenvectors ${\bf w}_{j}$ of the matrix ${\bf \tilde{S}}$.  
From here the $\jth{j}$ DMD basis function mode, or the approximate $\jth{j}$ eigenvector of the Koopman operator ${\bf A}$, can be shown to be 
\beq\la{equ:DMDBasisFunctionModes}
	{\bf \bfvarphi}_{j} = {\bf U}{\bf w}_{j},
\eeq
each of which can be put into the columns of a matrix $\bfPhi$~\cite{DMD1}.  
For the benefit of predicting the time dynamics, the DMD eigenvalues can be converted to Fourier modes by defining
\beq\la{equ:FourierFrequencies}
	\omega_{j} = \frac{\ln(\mu_{j})}{\D{t}}.
\eeq
The real part of $\omega_{j}$ regulates the growth or decay of the DMD basis function modes, while the imaginary part of $\omega_{j}$ drives oscillations in the DMD modes.  
Clearly, any unwarranted growth or decay in the Fourier modes will eventually limit the range in time that the model can credibly be predictive~\cite{DataBook}.  

Therefore, the DMD reconstruction of the data ${\bf x}_{\text{DMD}}$ at time $t$ for any time after the initial data vector ${\bf x}_{1}$ was collected ($t_{1} = 0$) is given by ${\bf x}_{\text{DMD}}(t)={\bf A}^{t}{\bf x}_{1}$.  
Using the approximate eigen-decomposition of the Koopman operator yields
\beq\la{equ:DMD}
	{\bf x}_{\text{DMD}}(t) = \sum_{j = 1}^{\ell} b_{j}\bfvarphi_{j}e^{\omega_{j} t} = \bfPhi\bfOmega^{t}{\bf b},
\eeq
where
\[
	\bfOmega = \usebox{\matOmega},
\]
and the vector ${\bf b}\approx\bfPhi^{-1}{\bf x}_{1}$ contains the initial amplitudes for the modes.  
At the time that the first snapshot of data was taken $(t_{1} = 0)$, equation \eqref{equ:DMD} reduces to
\[
	{\bf x}_{\text{DMD}}(t_{1}) = {\bf x}_{1} = \bfPhi{\bf b};
\]
hence, ${\bf b} = (\bfPhi^{*}\bfPhi)^{-1}\bfPhi^{*}{\bf x}_{1}$~\cite{DataBook}.  

Note that the DMD method demands that data is collected over evenly spaced time intervals $\D{t}$, but it cannot know for sure whether this was done in practice or not; thus it must deal with the data that it is given, assuming regular time intervals.  
This implies that there must be certain types of dynamics in time that the DMD method cannot model well even when the data is collected in periodic intervals.  
This is because there can be one-to-one mappings from certain types of dynamics taken at irregular time intervals to transformed dynamics taken at evenly spaced intervals, which can be accomplished through some kind of frame-of-reference transform.  

\subsection{Video Interpretation of the DMD Method}\la{subsect:VIDMDM}

 A video sequence offers an appropriate application for this DMD method because the frames of the video are, by nature, equally spaced in time, and the pixel data, collected in every snapshot, can readily be vectorized.  
Given $m$ frames to the video stream, the $n \times 1$ vectors ${\bf x}_{1}, {\bf x}_{2}, \ldots, {\bf x}_{m}$ can be extracted, which contain the pixel data of each frame; there being $n$ pixels in total per frame.  
The DMD method can attempt to reconstruct any given frame, or even possibly future frames, by calculating ${\bf x}_{\text{DMD}}(t)$ at the corresponding time $t$, as is described above in equation \eqref{equ:DMD}.  
The validity of the reconstruction depends on how well the specific video sequence meets the assumptions and criteria of the DMD method.  

In order to reconstruct the entire video, consider the $1 \times m$ time vector ${\bf t} = [t_{1} \ t_{2} \ \ldots \ t_{m}]$, which contains the times at which the frames were collected.  
If $t_{j} = j - 1 \ \forall j$, then time becomes equivalent to the frame count, where the first frame is labelled as 0 and the $\jth{m}$ frame is labelled as $m - 1$.  The video sequence ${\bf X}$ is reconstructed with the DMD technique as follows:
\beqs
	{\bf X}_{\text{DMD}} = \sum_{j = 1}^{\ell} b_{j}\bfvarphi_{j}e^{\omega_{j} {\bf t}} := \bfPhi\bfOmega^{{\bf t}}{\bf b}.
\eeqs
Notice that $\bfvarphi_{j}$ is a $n \times 1$ vector, which is multiplied by the $1 \times m$ vector ${\bf t}$, to produce the proper $n \times m$ video size.  
By the construction of the DMD methodology: ${\bf x}_{1} = \bfPhi{\bf b}$, which means that $\bfPhi{\bf b}$ renders the first frame of the video with a dimensionality reduction chosen through the parameter $\ell$.  
Thus, the diagonal matrix $\bfOmega^{t}$ dictates how that first frame gets altered over time in order to reconstruct the subsequent frames.  
It becomes apparent that any portion of the first video frame that does not change in time, or changes very slowly in time, must have an associated Fourier mode $(\omega_{j})$ that is located near the origin in complex space: $\|\omega_{j}\| \approx 0$.  
This fact becomes the key principle that makes possible the ability of the DMD method to separate background (approximate low-rank) information from foreground (approximate sparse) information.  

Assume that $\omega_{p}$, where $p \in \{1,2,\ldots,\ell\}$, satisfies $\|\omega_{p}\| \approx 0$, and that $\|\omega_{j}\| \ \forall \ j \neq p$ is bounded away from zero.  
Thus, 
\beq\la{equ:DMDTerms}
	{\bf X}_{\text{DMD}} = \underbrace{b_{p}\bfvarphi_{p}e^{\omega_{p} {\bf t}}}_{\text{Background Video}} + \underbrace{\sum_{j \neq p} b_{j}\bfvarphi_{j}e^{\omega_{j} {\bf t}}}_{\text{Foreground Video}}
\eeq
Assuming that ${\bf X} \in \R^{n \times m}$, then a proper DMD reconstruction should also produce ${\bf X}_{\text{DMD}} \in \R^{n \times m}$.  
However, each term of the DMD reconstruction is complex: $b_{j}\bfvarphi_{j}\exp\left(\omega_{j} {\bf t}\right) \in \C^{n \times m} \ \forall j$, though they sum to a real-valued matrix.  
This poses a problem when separating the DMD terms into approximate low-rank and sparse reconstructions because real-valued outputs are desired and knowing how to handle the complex elements can make a significant difference in the accuracy of the results.  
Consider calculating the DMD's approximate low-rank reconstruction according to
\[
	{\bf X}_{\text{DMD}}^{\text{Low-Rank}} = b_{p}\bfvarphi_{p}e^{\omega_{p} {\bf t}}.
\]
Since it should be true that
\[
	{\bf X} = {\bf X}_{\text{DMD}}^{\text{Low-Rank}} + {\bf X}_{\text{DMD}}^{\text{Sparse}},
\]
then the DMD's approximate sparse reconstruction,
\[
	{\bf X}_{\text{DMD}}^{\text{Sparse}} = \sum_{j \neq p} b_{j}\bfvarphi_{j}e^{\omega_{j} {\bf t}},
\]
can be calculated with real-valued elements only as follows\dots
\[
	{\bf X}_{\text{DMD}}^{\text{Sparse}} = {\bf X} - \Big|{\bf X}_{\text{DMD}}^{\text{Low-Rank}}\Big|,
\]
where $|\cdot|$ yields the modulus of each element within the matrix.  
However, this may result in ${\bf X}_{\text{DMD}}^{\text{Sparse}}$ having negative values in some of its elements, which would not make sense in terms of having negative pixel intensities.  
These residual negative values can be put into a $n \times m$ matrix ${\bf R}$ and then be added back into ${\bf X}_{\text{DMD}}^{\text{Low-Rank}}$ as follows:
\begin{align*}
	{\bf X}_{\text{DMD}}^{\text{Low-Rank}} & \leftarrow {\bf R} + \Big| {\bf X}_{\text{DMD}}^{\text{Low-Rank}} \Big| \\
	{\bf X}_{\text{DMD}}^{\text{Sparse}} & \leftarrow {\bf X}_{\text{DMD}}^{\text{Sparse}} - {\bf R}
\end{align*}
This way the magnitudes of the complex values from the DMD reconstruction are accounted for, while maintaining the important constraints that
\[
	{\bf X} = {\bf X}_{\text{DMD}}^{\text{Low-Rank}} + {\bf X}_{\text{DMD}}^{\text{Sparse}},
\]
so that none of the pixel intensities are below zero, and ensuring that the approximate low-rank and sparse DMD reconstructions are real-valued.  
This method seems to work well empirically.  

In terms of video streams, dimensionality reduction, done through the parameter $\ell$, has the effect of blurring the frames together through time.  
In this paper, $\ell$ is fixed to be the as large as possible, which is one less than the number of frames in the video sequence, or $m - 1$.  
Though, blurring frames together may be advantageous, as will be discussed later.

\section{DMD and video applications}\la{sect:SVA}

The computer vision application of separating background/foreground information in video sequences will now be considered using the DMD methodology.  

\subsection{Surveillance Video Application}\la{subsect:SVA}

Using the Advanced Video and Signal based Surveillance (AVSS) Datasets~\cite{AVSS}, specifically the ``Parked Vehicle - Hard'' and ``Abandoned Bag - Hard'' videos, the DMD separation procedure can be compared and contrasted against the RPCA procedure.  
The original videos are converted to grayscale and down-sampled in pixel resolution to $n = 120 \times 96 = 11520$, in order to make the computational memory requirements manageable for personal computers.  
Also, the introductory preambles to the surveillance videos, which constitute the first 351 frames of each video, are removed because they are irrelevant for the following illustrations.  

The video streams are broken into segments of $m = 30$ frames each, which are analyzed individually using both the RPCA and DMD methods.  
Frame numbers 500, 1000, and 2000 of the entire video stream are depicted in Figs.~\ref{fig:PVExamples} and \ref{fig:ABExamples}, along with their separation results for easy comparison.  
Although one has the option of manually tuning the regularization parameter of the RPCA method to best suite the given application, for fairness $\lambda$ is set at $\lambda = (\sqrt{n})^{-1} \approx 9.32 \cdot 10^{-3}$, as is suggested by Cand\`{e}s et al.~\cite{RPCA1} for creating a reliable automatic algorithm.  
Likewise, the dimensionality reduction parameter of the DMD method is held constant at $\ell = m - 1 = 29$ for the sake of producing the best results.  
For enhanced contrast and better visibility, the sparse results from both methods are artificially brightened by a factor of 10.  

Consider Fig.~\ref{fig:PVExamples} of the AVSS ``Parked Vehicle'' surveillance video, which generally shows various vehicles traveling along a road, with a traffic light (not visible) and a crosswalk (visible) near the bottom of the frame, and with an occasional vehicle parking along side the road.  
Sometimes, In the distance, moving vehicles become difficult to perceive with the naked-eye, due to limitations of the pixel resolution.  
For all three frames, the DMD method seems to eliminate more spurious pixels in its sparse results that may pertain to the background when compared to the RPCA's sparse results.  
However, one cannot be sure that these spurious pixels are truly erroneous.  
Note that some extraneous pixels in the sparse structure can be eliminated by applying a simple thresholding criterium based on pixel intensity.  

\begin{figure*}[t]
	\bc
   		\includegraphics[width=0.95\textwidth]{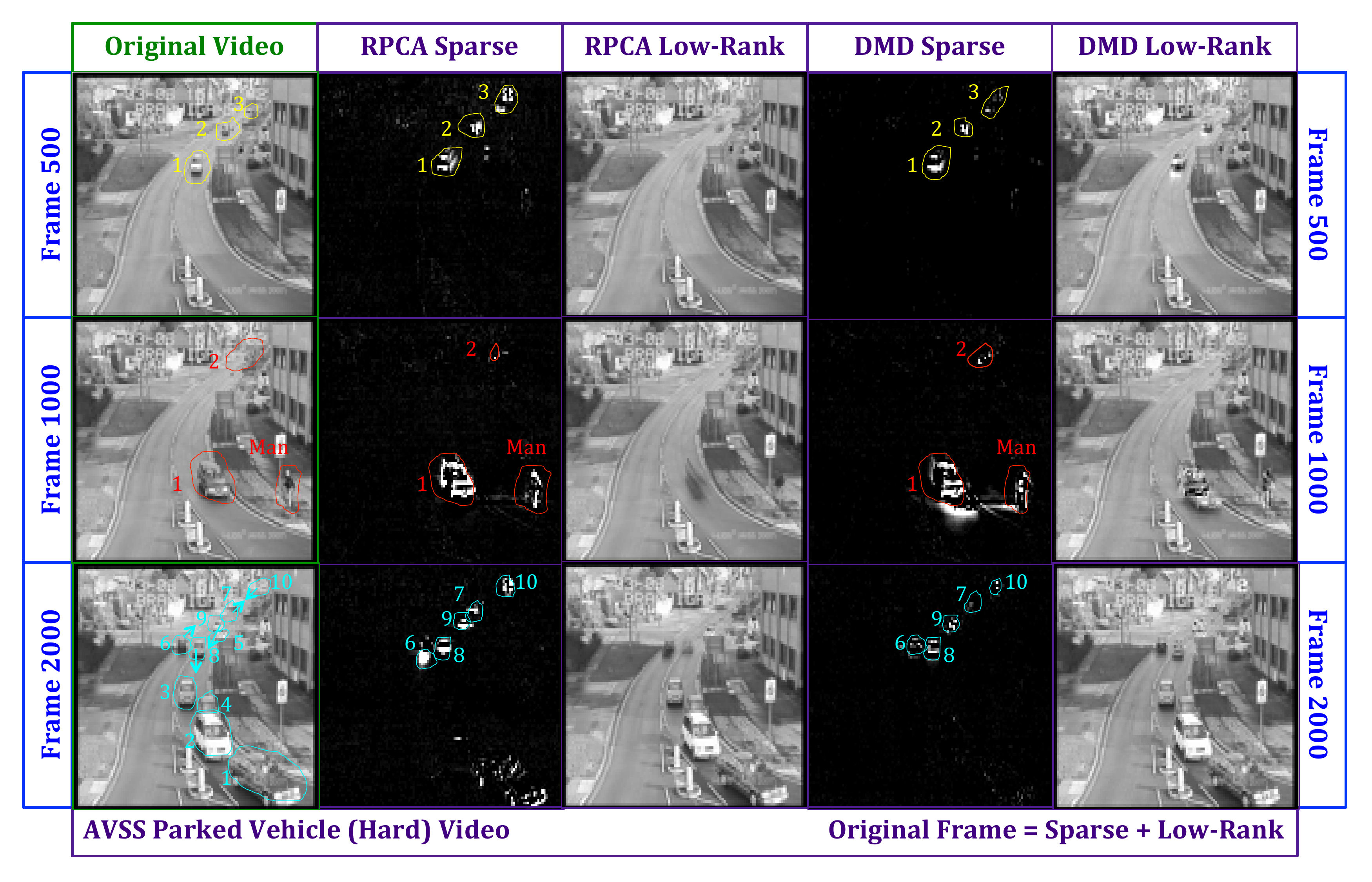}
	\ec
 	\caption{The DMD and RPCA background/foreground separation results are illustrated for 3 specific frames in the ``Parked Vehicle'' video.  
	The 30 frame video segment that contains frame $\#500$ has 3 vehicles driving in the same lane toward the camera.  
	The 30 frame video segment that contains frame $\#1000$ has a man stepping up to a crosswalk as a vehicle passes by, and a second car in the distance, barely perceptible, starts to come into view.  
	The 30 frame video segment that contains frame $\#2000$ has 3 vehicles stopped at a traffic light at the bottom of the frame with another 2 vehicles parked on the right side of the road, and 5 moving vehicles, 2 going into the distance and 3 coming toward the vehicles waiting at the light, the last vehicle being imperceptible to the eye at this pixel resolution: $n = 11520$.}
	\la{fig:PVExamples}
\end{figure*}

Consider Fig.~\ref{fig:ABExamples} of the AVSS ``Abandoned Bag'' surveillance video, which generally depicts people walking, standing, and sitting as trains come and go in a subway station.  
Shadows are relevant in this video because they do move with their respective foreground objects, and they do sometimes change the background significantly enough to be viewed as extensions of the moving objects themselves.  
Note that, for frames 500 and 1000, both methods struggle with the fact that between the numerous moving objects and their shadows, many of the pixels in the video change intensity at some point.  
Observe that objects in motion create movement trails that extrapolate the object's motion both forwards and backwards in time.  
When the object is dark, its corresponding movement trail is bright, and vice-versa.
The DMD method is generous in depicting these movement trails in its sparse results, while the RCPA method is more conservative.  

\begin{figure*}[t]
	\bc
   		\includegraphics[width=0.95\textwidth]{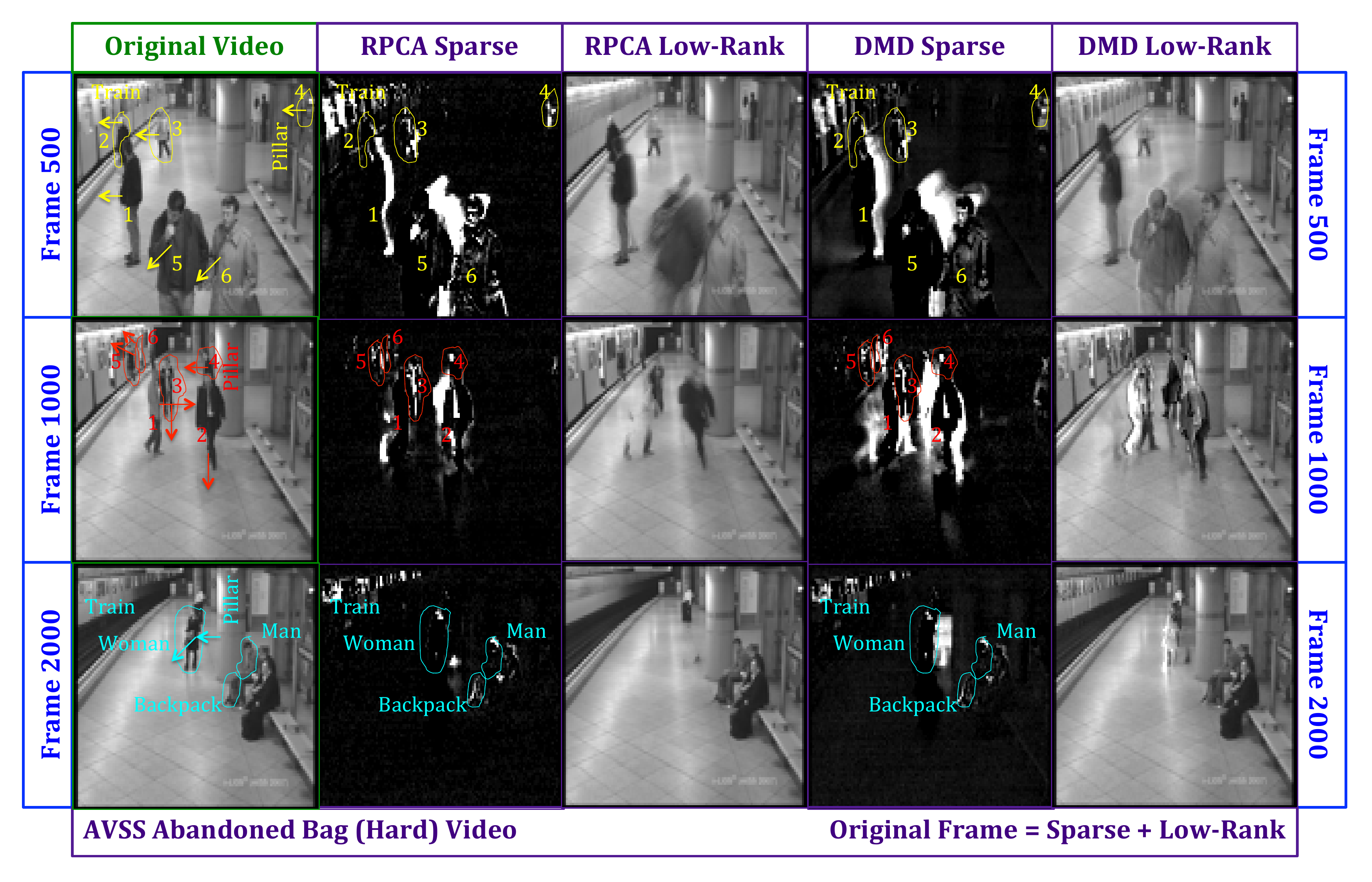}
	\ec
 	\caption{The DMD and RPCA background/foreground separation results are illustrated for 3 specific frames in the ``Abandoned Bag'' video.  
	The 30 frame video segment that contains frame $\#500$ has 3 people stepping slightly closer to an arriving train, and another person walks behind a support pillar, toward the train in the upper right area of the frame.  
	The two people closest to the camera walk toward the bottom of the frame.  
	The 30 frame video segment that contains frame $\#1000$ has 4 people walking in different directions in the middle of the frame, one of which comes out from behind a support pillar, while, farther down the platform, 2 people enter the train.  
	The 30 frame video segment that contains frame $\#2000$ has a woman walk out from behind a support pillar, moving to the left, and then turning somewhat toward the camera.  
	The train is moving and the man sitting closest to the support pillar adjusts his backpack.}
	\la{fig:ABExamples}
\end{figure*}

Figures~\ref{fig:PVExamples} and \ref{fig:ABExamples} seem to indicate that the RPCA and DMD methods have comparably good background/foreground separation results.  
The difference in the separation quality seems to depend on the specific situation, likely because the assumptions that determine when the RPCA and DMD methods will perform well are also different.  
Though, given the relatively consistent quality of the separation results, other factors, such as computational effort, become the distinguishing characteristics that determine which method is more suitable for the given application.  

\subsection{Timing Performance}\la{subsect:TP}

The real difference in effectiveness between the RPCA and DMD algorithms is found in the amount of computational time needed to complete the background/foreground separation.  
Again, consider the AVSS Datasets, with the two videos used previously: ``Parked Vehicle'' and ``Abandoned Bag''.  
For a timing performance experiment, consider having these videos down-sampled to various pixel resolutions $n$ and separated into various video segment sizes $m$.  
Averaging the computational time for either fixed numbers of pixels or fixed video segment sizes, and using both the RPCA method and the DMD method on both videos, Fig.~\ref{fig:DMDTimingPerformance} was produced.  
Both the exact ALM and the faster inexact ALM convex optimization routines were used for solving the PCP problem~\eqref{equ:PCP} of the RPCA method.  

\begin{figure*}[t]
	\bc
		\subfloat[][Fixed pixel resolution.]{\la{fig:TimingVariedFrames}\includegraphics[width=0.95\textwidth]{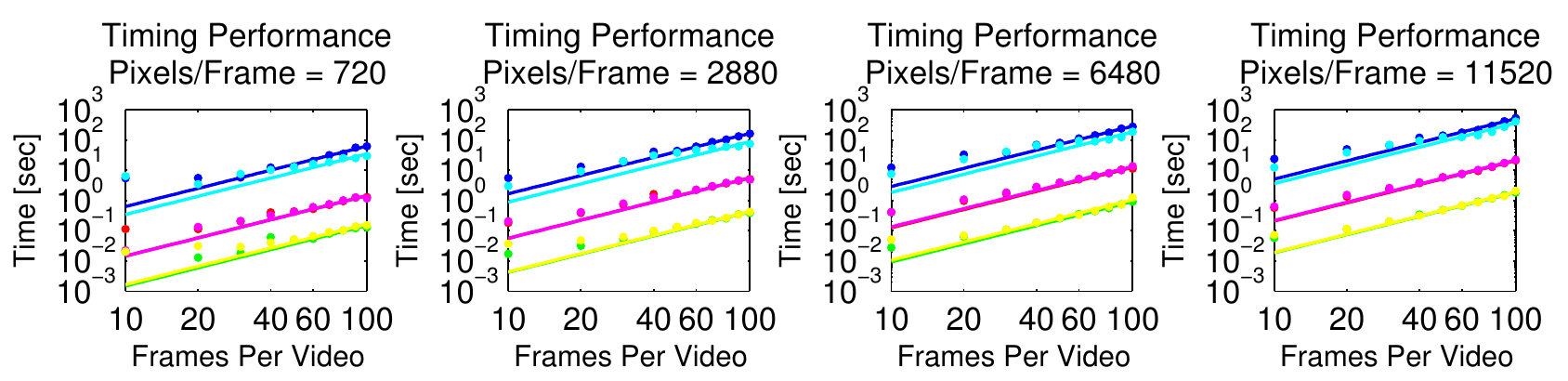}} \\
		\subfloat[][Fixed video segment size.]{\la{fig:TimingVariedPixels}\includegraphics[width=0.95\textwidth]{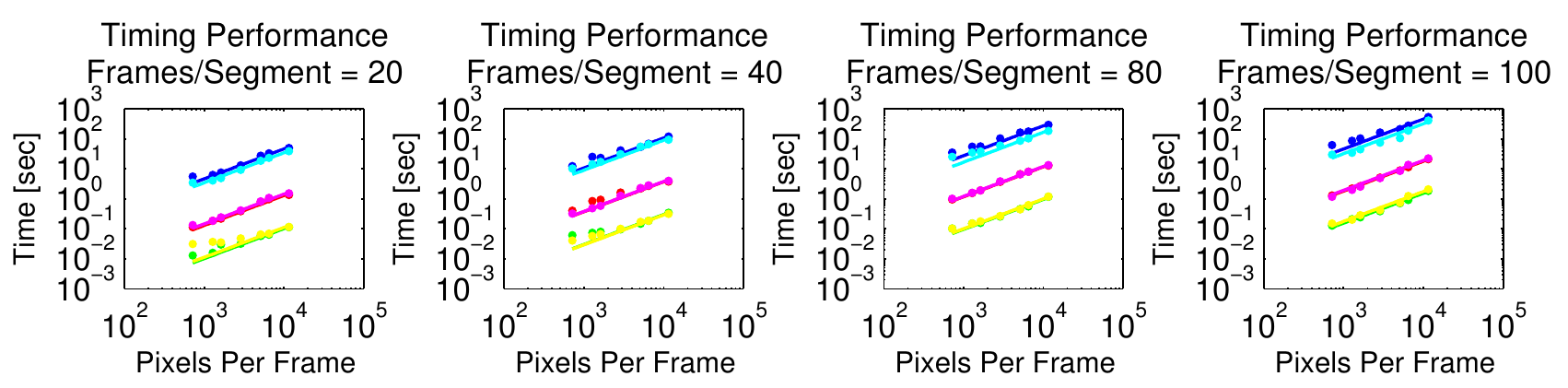}}
	\ec
 	\caption{DMD timing performance on a 1.86 GHz Intel Core 2 Duo processor using Matlab.  
	For the ``Abandoned Bag'' and ``Parked Vehicle'' videos, the computational times of the DMD (green \& yellow, respectively), inexact ALM RPCA (red \& magenta, respectively), and exact ALM RPCA (blue \& cyan, respectively)  background/foreground separation methods are graphed on a logarithmic scale.  
	Figure (a) holds the numbers of pixels per frame fixed at $\{720,2880,6480,11520\}$.  
	This timing data fits reasonably well with a quadratic fit: $t = cs^{2}$, where $t$ is the computational time, $c \in \R$, and $s$ is the video segment size (number of frames per video segment).  
	Figure (b) holds the numbers of frames per video segment fixed at $\{20,40,80,100\}$.  
	This timing data fits reasonably well with a linear fit: $t = cs$, where $t$ is the computational time, $c \in \R$, and $s$ is the frame resolution size (number of pixels per frame).}
	\la{fig:DMDTimingPerformance}
\end{figure*}

In Fig.~\ref{fig:DMDTimingPerformance}, the empirical computational times are plotted on a logarithmic scale, along with best-fit curves found by the linear least squares method.  
It is clear that the DMD method is about $2 - 3$ orders of magnitude faster than its RPCA method counterpart using the exact ALM optimization procedure, and is about 1 order of magnitude faster when the inexact ALM optimization procedure is employed.  
In fact, given that many cameras operate at a rate of about $20 - 30$ frames per second and that the DMD method can be completed for those video segment sizes in about $0.1 - 0.01$ seconds for high and low resolution images, respectively, then real-time, on-line data processing is possible, even without downsampling.  

\subsection{DMD and Dimensionality Reduction}\la{subsect:PT}

As was mentioned previously, there is an option for implementing dimensionality reduction in the DMD algorithm.  
In order to portray the effect that the dimensionality reduction parameter $\ell$ has on the DMD separation process, consider the foreground DMD results represented in Fig.~\ref{fig:DimensionalityTest}.  

\begin{figure*}[t]
	\bc
   		\includegraphics[width=0.95\textwidth]{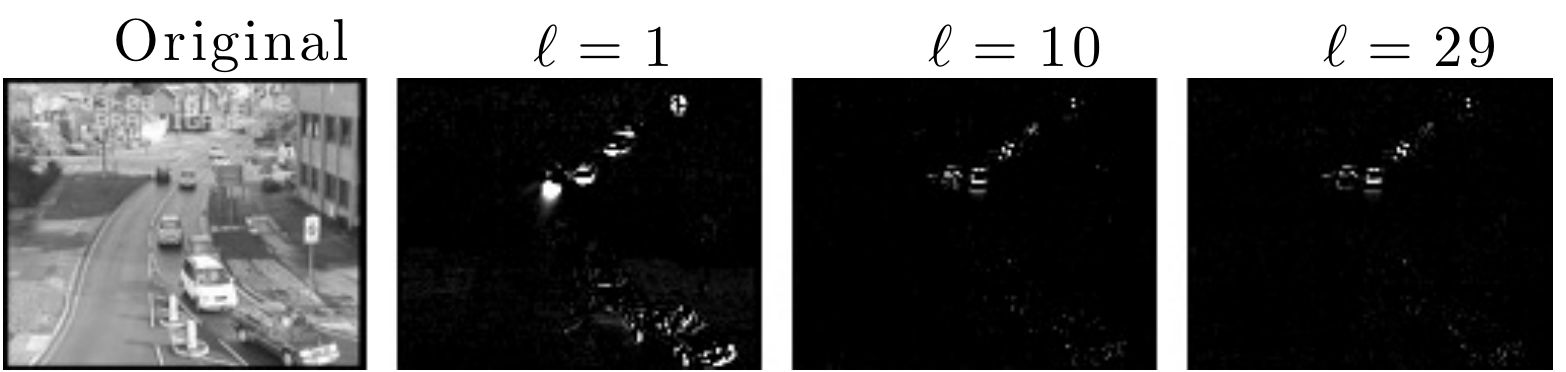}
	\ec
 	\caption{The effects of the parameter $\ell$ that controls the dimensionality reduction of the DMD process is depicted.  
		The original frame, and the DMD sparse reconstructions of that frame, artificially brightened by 10 times the true intensity, are illustrated for frame $\#2000$ of the ``Parked Vehicle'' video~\cite{AVSS} using a 30 frame video segment (See Fig.~\ref{fig:PVExamples}).  
		Notice that the sparsity increases with the parameter $\ell$ because dimensionality reduction has the effect of blurring the motion between frames, and thus smearing the motion out over a greater number of pixels in any given frame.  
		Also, note that even under tremendous dimensionality reduction $(\ell = 1)$ the background/foreground separation is still accomplished reasonably well.  
		Compare this to changing the regularization parameter $\lambda$ in the RPCA method of Fig.~\ref{fig:LambdaTest}, where $\ell = 29$ for DMD method in that figure.}
	\la{fig:DimensionalityTest}
\end{figure*}
  
The surprising result depicted in this figure is the quality of the low-rank and sparse separation even when $\ell$ is set as low as 1; compare this to when $\ell = m -1$.  
There is some indication here that the low dimensional results may even be better than the high dimensional results; this will need further exploration in the future.  
Dimensionality reduction seems to have the effect of blurring the video frames together.  
It is possible that the good results of $\ell = 1$ are in part due to the fact that the vehicles in the video of Fig.~\ref{fig:DimensionalityTest} did not move very far within the frame, and so the smearing of pixels over the places of movement is very localized.  
This dimensionality reduction constitutes a {\em time-saving} procedure that can potentially reduce the video separation costs even further.  

\subsection{DMD Error Analysis}\la{subsect:DEA}

In order to accurately measure how well the DMD method captures actual foreground movement, and not the stationary background, it is helpful to have an artificially constructed video where the true background and foreground are known. 
The error could then be measured precisely between the actual and DMD reconstructed backgrounds.  
Moreover, the quality of reconstruction can be also compared with the RPCA technique.  

\begin{figure*}[t]
	\bc
		\subfloat[][Constructed Video DMD Separation]{\la{fig:constructed}\includegraphics[width=0.7\textwidth]{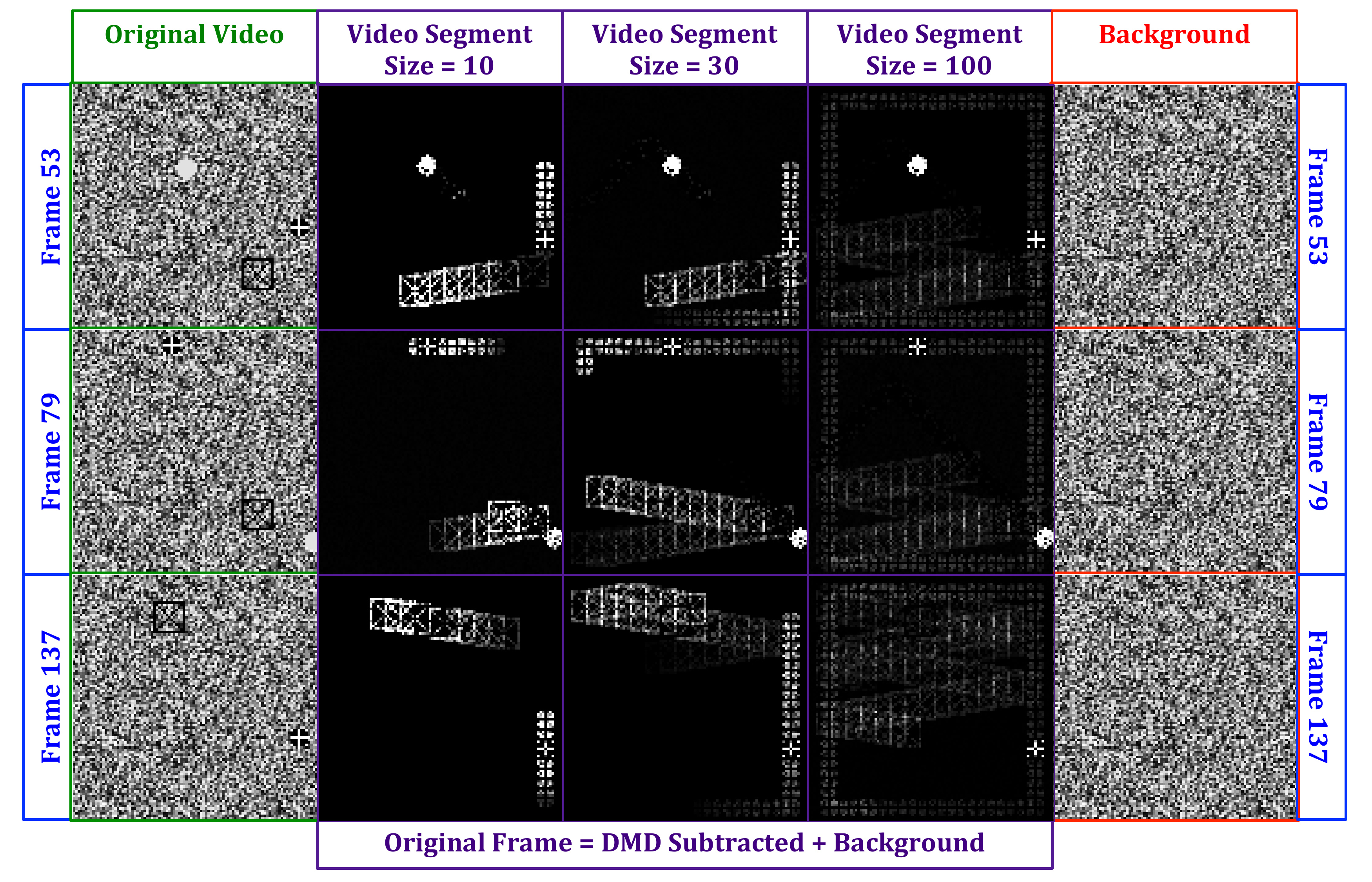}} \\
		\subfloat[][Error Analysis]{\la{fig:error}\includegraphics[width=0.32\linewidth]{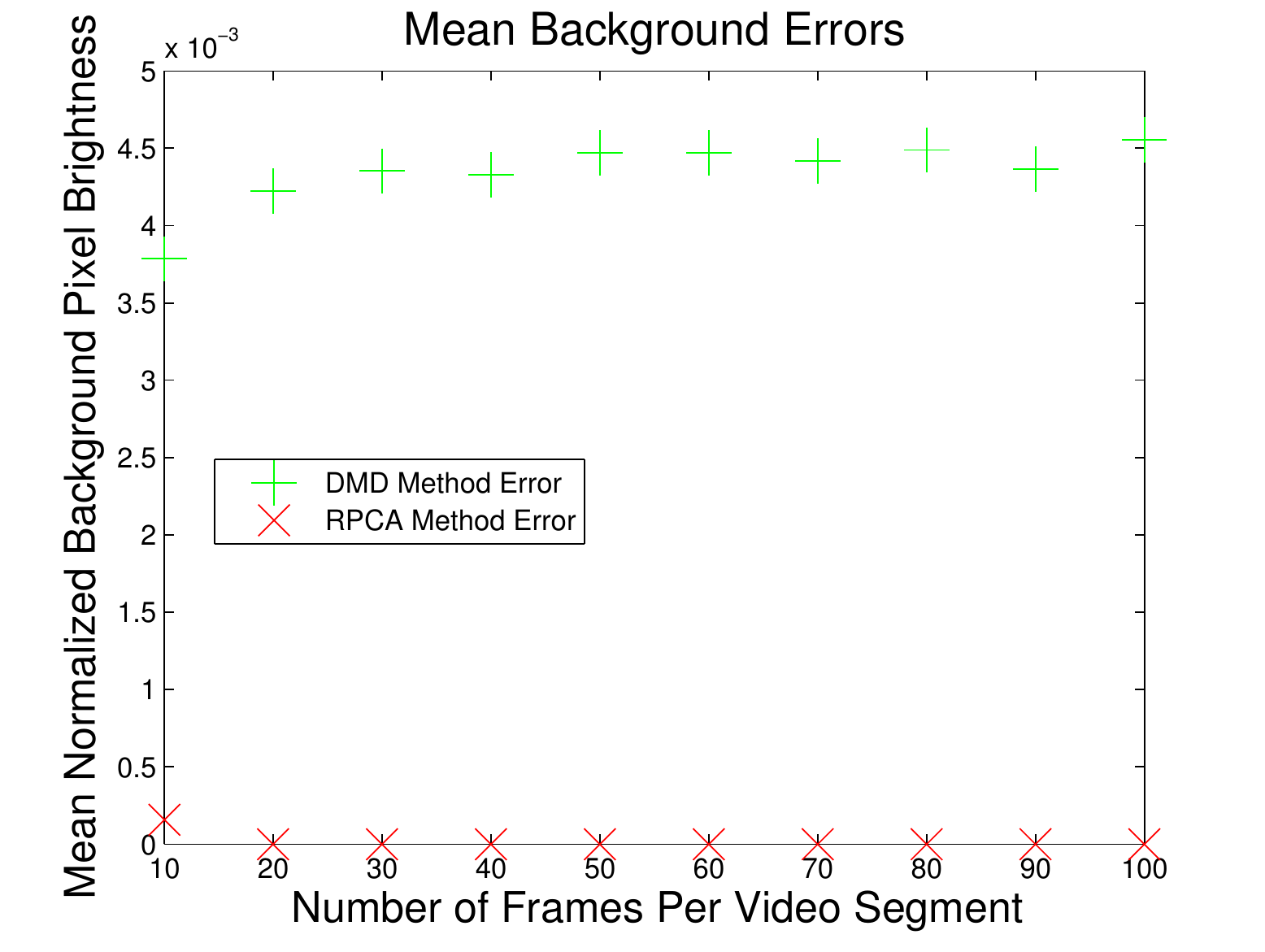}} \quad
		\subfloat[][Timing Analysis]{\la{fig:timing}\includegraphics[width=0.32\textwidth]{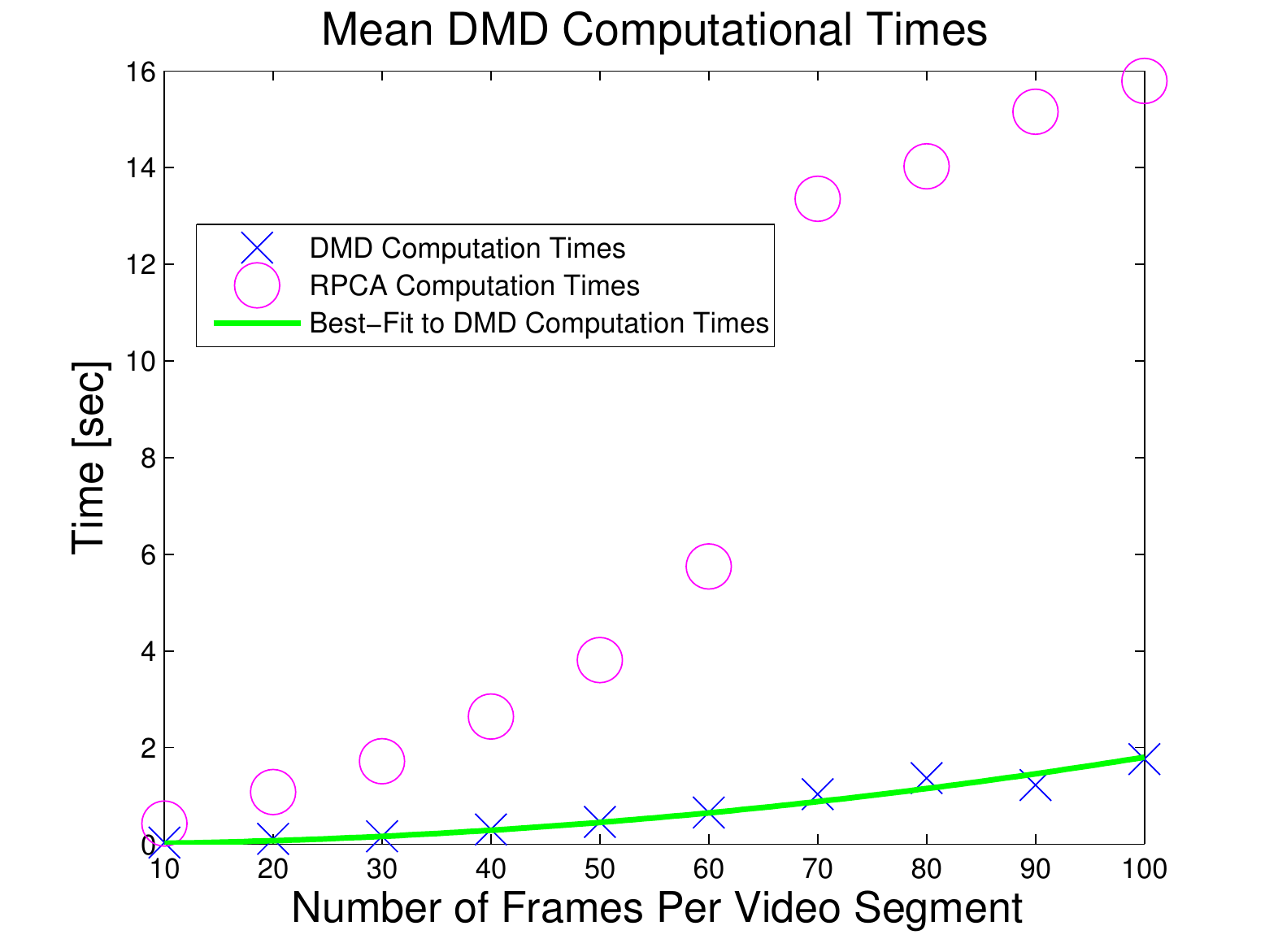}}
	\ec
 	\caption{The DMD background/foreground separation results for the 300 frame video used for an error analysis are presented here.  
	In the top figure (a), the left column shows the original frames with the random background noise that is fixed in time.  
	The middle three columns show the sparse DMD results using 10, 30, and 100 frames per video segment, which are artificially brightened by a factor of 10 in order to increase the contrast and visibility of the results against a black background, and the right column shows the true background noise that was added to each frame.  
	The erroneous pixels contribute to the mean pixel intensity error of the background.  
	The mean pixel intensity error for the entire video at each video segment size is plotted in the bottom left figure (b) on a normalized intensity scale where 1 is pure white and 0 is pure black, the target background color.  
	The mean computational times $t$, using Matlab on a 1.86 GHz Intel Core 2 Duo processor, as a function of the video segment sizes $s$ is plotted in the bottom right figure (c); yielding a quadratic best-fit: $t = 1.52 \cdot 10^{-4}s^{2}$, $R^{2} = 0.983$.}
	\la{fig:ErrorAnalysis}
\end{figure*}

One such video was constructed for calculating the error in the DMD low-rank/sparse separation method with 300 frames in total, each being $100 \times 100$ pixels.  
The background to the video is produced by a uniform random, grayscale intensity field, ranging form pure black to pure white.  
The background remains constant throughout the entire video.  

This video contains three moving objects: (1) a black square, $7 \times 7$ pixels in size, with four white pixels in each corner and a white "plus" sign centered in the middle, (2) a light gray circle with a diameter of 9 pixels, and (3) a transparent square, $13 \times 13$ pixels in size, lined by black pixels and with a black ``X'' centered within.   
Object (1) revolves, at a constant rate of 4 pixels per frame, counter-clockwise around the inside edges of the frame starting at the top left corner of the frame for the entire duration of the video.  
Object (2) enters the video on frame 25 on the left side of the frame, moving up and to the left at a constant rate of 2 pixels per frame in both directions.  
This object reflects downward and eventually leaves the frame after bouncing off of an imaginary wall located a fifth of the way down from the top of the frame.  
Object (3) enters the video on frame on frame 50 on the right side of the frame, moving down and to the left.  
Once this object enters the frame it stays inside the frame for rest of the video, reflecting off of the edges of the frame.  
Object (3) maintains a constant vertical velocity of 1 pixel per frame, and a horizontal velocity of 6 pixels per frame.  
At various times these objects overlap one another, and the precedence is that object (1) is drawn first, then object (2), then object (3), burying the previously drawn objects underneath the newly drawn object.  

The foreground results of the DMD method applied to this video are illustrated in Fig.~\ref{fig:constructed} for frame numbers 53, 79, and 137.    
Note that object (3) cannot be seen in the results because it is pure black.  
This exemplifies a limitation to the DMD method, in that when there is an object of low pixel intensity, it may be difficult to distinguish it from the zero entries that naturally occur in a sparse matrix.  
Note that the spurious pixels are both residuals of previous frames and projections of where the objects will be moving in future frames, and have inverted pixel intensities compared to their corresponding objects' intensities.  
This is much like what is seen in the foreground DMD results of frames $\#1000$ and $\#2000$ of the ``Abandoned Bag'' video Fig.~\ref{fig:ABExamples}, where the walking man (frame $\#1000$) and walking woman (frame $\#2000$) have obvious, erroneous movement trails.

The mean, normalized pixel intensity error for this constructed video is shown in Fig.~\ref{fig:error}.  
This error is calculated on the first $m - 1$ frames of the video segments of length $m$ because the last frame is where the DMD procedure accumulates its error.  
Note that the background to the sparse DMD reconstructed videos should be pure black, and on an intensity scale of 0 to 1, the average sparse background pixel intensity is on the order of $10^{-3}$, compared to the true average background intensity of nearly identically 0.5.  
Recall that, in some cases, thresholding techniques can eliminate spurious background pixels from the foreground results, and improve the measured error.  

One might have expected that the DMD algorithm errors would have decreased as the video segment sizes increased, due to the fact that having more frames means that the DMD method has more information to work with.  
However, because of this anomaly with black objects leaving white movement trails, there are extraneous pixels for every video segment size.  
As the videos get longer, these erroneous movement trails also get longer, projecting both into the past and future increasing the amount of error in proportion to the increase in number of frames per video segment.  
However, the longer videos do produce less bright pixel intensity trails, which helps reduce the error, which is consistent with the idea that they should be able to make more use of the extra information that they have.  

The RPCA reconstructed foreground error results are also presented for comparison, where the suggested $\lambda = (\sqrt{n})^{-1} = 0.01$ is used.  
In this case, the RPCA perfectly reconstructs every video for segment sizes greater than 10 frames.  

Figure~\ref{fig:timing} reconfirms the quadratic growth in computational times that the DMD and RPCA scheme experience as the video segment sizes are increased.  
This quadratic growth trails off slightly for very small segment sizes, likely due to inherent programming processing times that do not scale with data size.  
For segment size 10, there is, of course, the option of retuning the regularization parameter $\lambda$ in RPCA.  However, in most cases where the true solution is unknown, this must be done manually by time-consuming reruns the RPCA algorithm.  

\begin{figure}[t!]
	\bc
   		\includegraphics[scale=0.5]{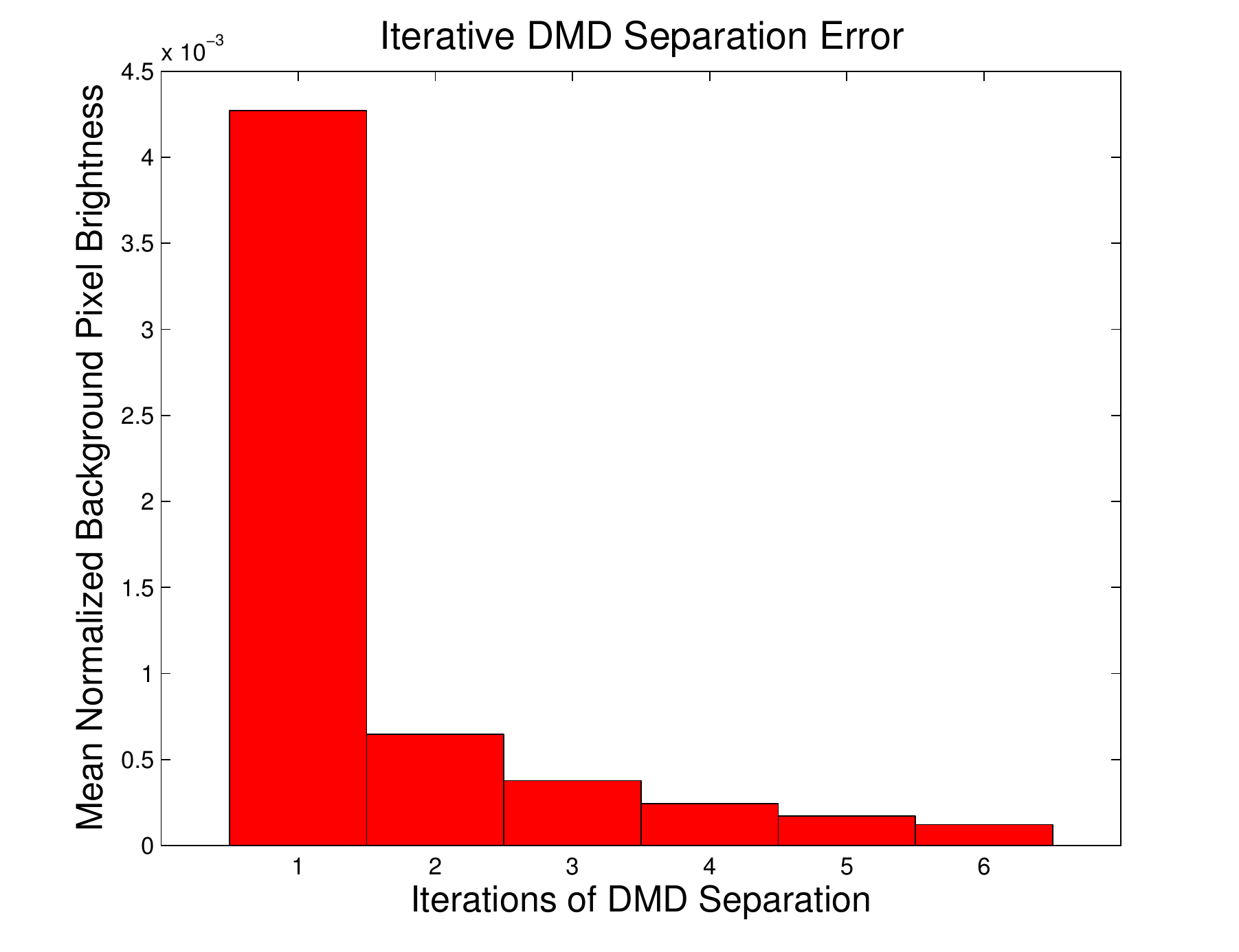}
	\ec
 	\caption[Iterative DMD Error]{In this figure, the mean normalized background pixel intensity, a measure of DMD separation error, is shown to decrease as the DMD separation method is applied iteratively to the previous iteration's sparse component result.  
	The first error depicted above $(i = 1)$ corresponds to first separation of the original video data, and the rest of the errors correspond to the iterations of consecutively applying the DMD method $(i = 2, 3, \dots)$.}
	\la{fig:IterativeDMDError}
\end{figure}

Finally, it should be noted that as object size increases relative to the frame size, the DMD reconstruction error also increases.  
Likewise, and not surprisingly, the DMD reconstruction error increases if less frames are used to show the same object movement.  
Not all types of object movement are accurately reconstructed with the DMD method; however, it seems that high-speeds and acceleration can still be handled fairly well given enough snapshots to capture the movement.   
Objects that were once moving and then stop is an event that is not well reconstructed by the DMD method.  

\subsection{Iterative Application of DMD}

One of the advantages of the RPCA method that is illustrated
in the preceding subsection is that it can perform exact low/rank
and sparse decompositions.  Thus the error can be reduced to zero.
No such guarantees are available for the DMD algorithm. However,
we highlight an interesting observation concerning the iterative use
of DMD.   Figure~\ref{fig:IterativeDMDError} demonstrates the fact that the DMD method can be applied iteratively in order to empirically converge toward the true low-rank and sparse components of the given data.  The successive DMD iterations are applied to the approximate sparse structure, creating a better approximate sparse structure and adding more data back into the low-rank structure.  
In the future, finding the exact rate of convergence and understanding the theoretical basis for this convergence will be important steps toward establishing an analytical connection between the RPCA and DMD separation algorithms.

\section{Conclusions and Outlook}

Overall it has been demonstrated that the method of dynamic mode decomposition, typically used for evaluating the dynamics of complex systems, can be used for background/foreground separation in videos with visually appealing results and excellent computational efficiency.  
The separation results produced by the DMD method are on par with the quality of separation achieved with the RPCA method for realistic video scenarios.  
However, the results are achieved orders of magnitude faster.  
Indeed, we demonstrate that DMD is viable as a real-time solution to foreground/background video separation tasks even with laptop-level computing platforms.

As with any separation method, including RPCA and DMD, the burden of working with too much data, i.e. high-resolution images and/or many frames per video segment, can be problematic because of reduced computational speeds and limited memory sizes.  
Nonetheless, the DMD algorithm has shown itself to be robust and efficient enough to produce attractive results in times well below the normal frame acquisition rate of most cameras, allowing for higher pixel resolutions and video segment sizes to be used.  
For real-time video applications, it makes sense to break the continuous video stream into segments large enough to ensure that there is enough information to complete an adequate background/foreground separation, but small enough to keep the processing times smaller than the data acquisition times.  
Additionally, moving objects that turn, stop, and/or accelerate are better handled by the DMD procedure as individual actions, than in one large video segment.

Finally, unlike RPCA, the analysis presented here is a simple formal procedure that is presented without proof of convergence to low-rank and sparse structures.  
Although the results are very promising in terms of both quality and exceptional computational costs, future work will focus on making rigorous the results under suitable conditions.  
Potentially, one could use the DMD technique for generating initial conditions for the convex optimization problem of RPCA, thus combining the power and speed of the method with the exact reconstruction ability of RPCA.  
It is also intriguing to consider the possibility of connecting the DMD methodology to the more standard $L^{1}$ convex optimization problem of RPCA and the matrix completion problem; perhaps allowing for even further improvement in background separation methods in terms of both quality and speed.  


\section*{Acknowledgements}  
J. N. Kutz acknowledges extensive and extremely informative conversations about DMD, sparsity, and its various implementations with Bingni Brunton, Steven Brunton, Mark Luchtenberg, Joshua Proctor, Clancy Rowley, and Jonathan Tu.  

{\small
\bibliographystyle{is-unsrt}
\bibliography{BackSubtractbib}
}

\end{document}